\documentclass[%
 reprint,
preprintnumbers,
 amsmath,amssymb,
 aps,
]{revtex4-2}

\usepackage{graphicx}
\usepackage[latin1]{inputenc}

\usepackage[nokeyprefix]{refstyle}
\usepackage{varioref}
\usepackage{xr-hyper}
\usepackage{hyperref}
\usepackage[dvipsnames]{xcolor}

\usepackage{amsmath}	
\usepackage{amssymb}	
\usepackage{xcolor}

\newcommand*{\bv}[1]{\mathbf{#1}}	

\newcommand{\nparams}{D}

\newcommand{\K}{K}

\newcommand{\xVec}{\mathbf{x}}

\newcommand{\Xt}{{\bv X}}
\newcommand{\Yt}{{\bv Y}}

\newcommand{\Rt}{{\bv R}}
\newcommand{\h}{{\bv h}}

\newcommand{\Tf}{{\bv T}}
\newcommand{\If}{{\bv I}}

\newcommand{\Y}{{\bv Y}}

\newcommand{\y}{{\bv y}}
\renewcommand{\h}{{\bv h}}

\renewcommand{\r}{{{r}}}
\newcommand{\noise}{\eta}

\newcommand{\transpose}{\top}

\newcommand{\pare}[1]{ \left(#1\right)}
\newcommand{\change}[1]{{\color{Black} #1}}

\DeclareMathOperator{\thetavec}{\boldsymbol \theta}
\DeclareMathOperator{\psivec}{\boldsymbol \psi}
\DeclareMathOperator{\z}{\mathbf{z}}

\definecolor{burgundy}{rgb}{0.5, 0.0, 0.13}

\newcommand{\affilPasteur}{Decision and Bayesian Computation, USR 3756 (C3BI/DBC) $\&$ Neuroscience department CNRS UMR 3751, Institut Pasteur, Universit\'e de Paris, CNRS, Paris, France}

\newcommand{\affilSanofi}{Histopathology and Bio-Imaging Group, Sanofi, R\&D, Vitry-Sur-Seine, France}

\begin{document}


\title{Variational inference of fractional Brownian motion with linear computational complexity}

\author{Hippolyte Verdier}
\email{hverdier@pasteur.fr}
\affiliation{\affilPasteur}
\affiliation{\affilSanofi}
\author{Fran\c{c}ois laurent}
\affiliation{\affilPasteur}
\author{Alhassan Cass\'e}
\affiliation{\affilSanofi}
\author{Christian L. Vestergaard}
\affiliation{\affilPasteur}
\author{Jean-Baptiste Masson}
\email{jbmasson@pasteur.fr}
\affiliation{\affilPasteur}

\begin{abstract}

We introduce a simulation-based, amortised Bayesian inference scheme to infer the parameters of random walks. Our approach learns the posterior distribution of the walks' parameters with a likelihood-free method. In the first step a graph neural network is trained on simulated data to learn optimised low-dimensional summary statistics of the random walk. In the second step an invertible neural network generates the posterior distribution of the parameters from the learnt summary statistics using variational inference. We apply our method to infer the parameters of the fractional Brownian motion model from single trajectories.
The computational complexity of the amortised inference procedure scales linearly with trajectory length, and its precision scales similarly to the Cram\'er-Rao bound over a wide range of lengths. The approach is robust to positional noise, and generalises to trajectories longer than those seen during training. Finally, we adapt this scheme to show that a finite decorrelation time in the environment can furthermore be inferred from individual trajectories.
 
\end{abstract}





\maketitle

\section{Introduction}

Fractional Brownian motion (fBm) ~\cite{Gardiner,Mandelbrot1968} is a paradigmatic model of anomalous transport. It is a non-Markovian Gaussian process characterised by stationary increments and long temporal correlations in the noise driving the process. It allows capturing long-range temporal correlations in the dynamics of a walker or its environment, and it is a model of choice to describe a multitude of dynamic processes in numerous scientific fields~\cite{Bouchaud1990,Leland1994,Cutland1995,Kukla1996,Decreusefond1998,  bouchaud2003theory, Weber2010,Dubbeldam2011,  Jeon2011, Walter2012, Ernst2012, Rostek2013, Palyulin2014,  Javer2014, Han2020, Wang2020, Gherardi2017,Arutkin2020, Yu2021}.
Following the classification given in~\cite{Metzler2014} of the three main sources of anomalous diffusion, the anomalous dynamics of fBm stems from the statistical dependency of the displacements at all time scales.  
Since fBm is a Gaussian process, it admits an analytical expression of the joint likelihood of the recorded signal. 
It is thus an ideal model to investigate the performance of approximate schemes to infer anomalous diffusion, such as variational inference or machine-learning-based approaches, since it allows direct comparison to statistically optimal exact inference.

The position of a random walker undergoing fBm is described by a Langevin equation~\cite{Gardiner} of the form
\begin{equation}
    \frac{d\r\pare{t}}{dt} = \sqrt{\K_{\alpha}}\, \noise\pare{t} ,
\end{equation}
where $\noise$ is a zero-mean Gaussian noise process with covariance  $\langle\noise \pare{t_{1}}\noise \pare{t_{2}}\rangle= \alpha\pare{\alpha-1}|t_{1}-t_{2}|^{\alpha - 2}$ and $\K_{\alpha}$ is a generalised diffusion constant that sets the scale of the process. 
fBm is self-similar and ergodic~\cite{Deng2009,Burov2011}. 
However, it has been shown to exhibit transient non-ergodic behaviour when confined~\cite{Burov2011,Jeon2012} and it is worth noting that the ergodic regime is witnessed only after a long transient passage exhibiting non-ergodic properties~\cite{Geneston2016,Burov2011,LochOlszewska2016,Kursawe2013}. The noise $\noise$ is negatively correlated in the subdiffusion regime ($\alpha < 1$), while it is positively correlated in the super-diffusion regime ($\alpha > 1$). 

Methods for estimating a random walk's parameters can roughly be divided into two types: heuristic approaches using features extracted from the trajectories~\cite{Meroz2015,Kosztoowicz2005,Han2020,Sanders2012,andi2020}, 
and likelihood-based (e.g., Bayesian) approaches~\cite{Lysy2016,Krog2018,Koo2016,Thapa2018}.
Each has its strengths and weaknesses.
Likelihood-based approaches are provably asymptotically optimal, but they are often computationally intensive and are only applicable to random walk models that have a tractable likelihood.
Feature-based approaches are typically computationally cheaper, and they can be applied to a much larger range of models since they do not require a tractable likelihood. 
However, they are generally not statistically efficient, are prone to bias when used on experimental data and their precision can be difficult to evaluate.
It is worth noting the rapid progress of machine learning based approaches~\cite{andi2020,munoz-gil_objective_2021}, which fall in the category of feature-based approaches, and which allow to learn high quality features to perform both parameter estimation and model classification. 
While such machine learning approaches generally outperform handcrafted features on numerically generated data, it remains difficult to evaluate their actual performance and robustness on empirical data.

\begin{figure*}[!tp]
    \centering
    \includegraphics[width=0.7\textwidth]{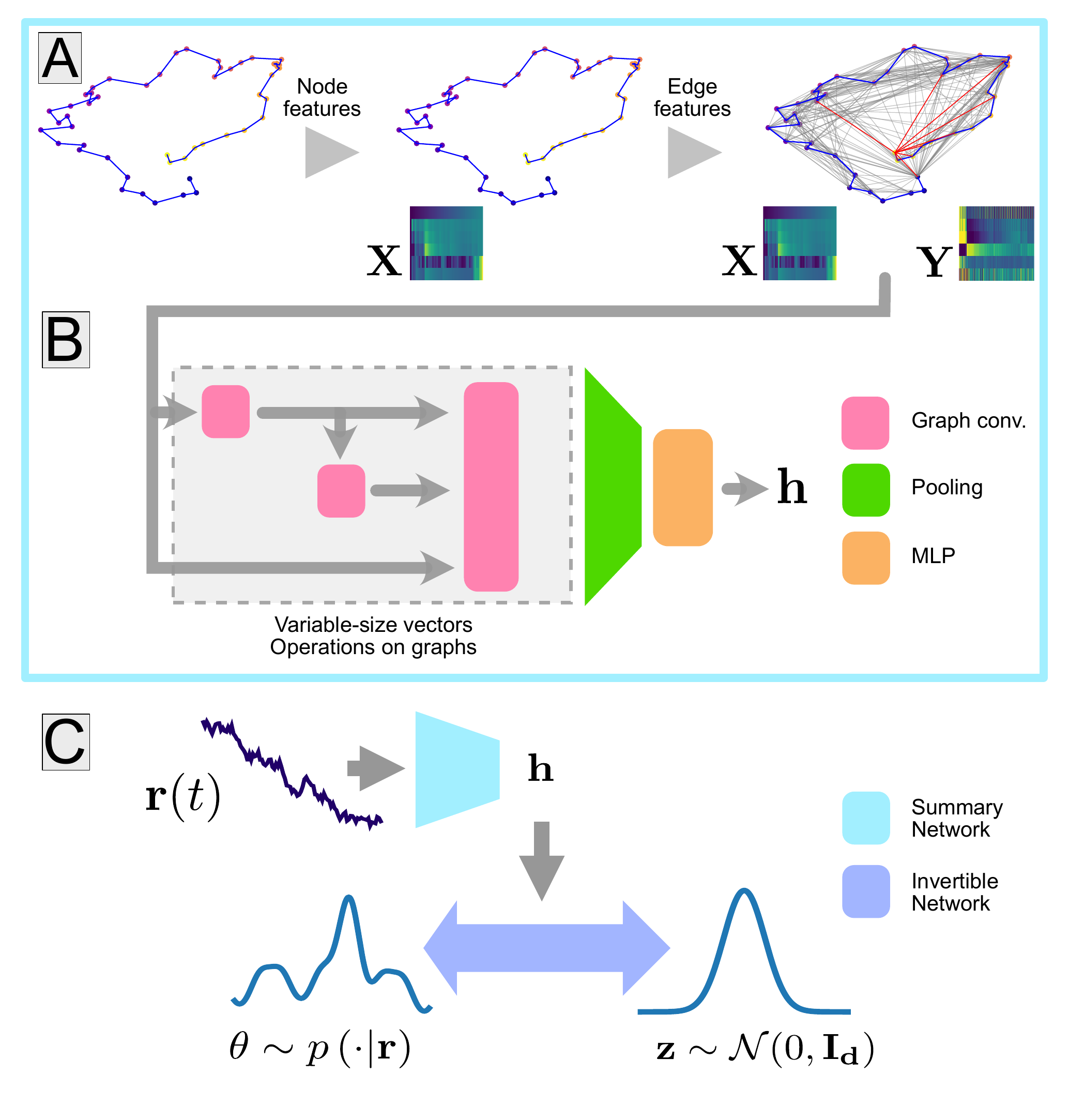}
    \caption{\textbf{Model Architecture.} 
    A: Construction of a graph from a single trajectory (left). Positions, colored according to time, are treated as nodes for which features $\mathbf{X}$ are computed (middle). Nodes are then connected by edges (grey lines) following a given wiring scheme, and edge features $\mathbf{Y}$ are computed. Edges terminating at the trajectory's last point are shown in red. Feature matrices for both nodes (X) and edges (Y) are depicted in small insets with color coded values.
    B: Summary network extracting information from the trajectory's graph. It consists in several graph convolution layers (in purple), a pooling layer (green) and a multi-layer perceptron (orange). The vector of statistics it outputs is indicated by \textbf{h}.
    C: General structure of the model, with the summary network (light blue) extracting from the trajectory $\textbf{r}(t)$ the summary statistics vector \textbf{h}, which in turn parameterises the invertible network. During the training phase, the invertible network is used from left to right (i.e., from the parameter's manifold to an easily sampled one), and in inference mode it is used from right to left.}
    \label{architecture-training}
\end{figure*}

Here, we develop an amortised \change{variational} inference approach to estimate the parameters of a fBm from a single recorded trajectory. \change{The objective of this paper is three-fold : 
i) we explored the possibility, via amortisation, to reduce the \change{marginal} computational cost of inference from quadratic to linear in the trajectory length, ii) we compared the performance of a such estimator with the Cramer-Rao bound and we assessed information retrieval by a linear computational scheme and iii) we investigated the possibility of using this variational approach to infer the posterior distribution of a finite decorrelation time of the walker's dynamics. }

The inference scheme relies on a graph neural network (GNN) trained on simulations of fBM realisations, which encodes a vector of  summary features for each trajectory. 
The encoder's architecture allows it to naturally capture long-range interactions while retaining a linear scaling of the computational complexity with the length of the trajectories.  It is associated with an invertible network, which generates the posterior distribution from the summary features using a variational objective. We applied the method to the fBm model because the tractability of its likelihood allows us to compute the Cram\'er-Rao bound which provides a lower bound on the variance of unbiased estimators. 
We show that the approach attains near-optimal performance as compared to exact likelihood-based inference and to the Cram\'er-Rao bound.
We furthermore discuss the latent space structure learned by the summary network and its ability to encode physical properties. 
We test the applicability of the approach to trajectories corrupted by positional noise and its potential to generalise to trajectories that are longer than those seen during training. 
Finally, we extend the inference procedure to capture a finite decorrelation time in the dynamics which may typically arise in physical environments.


\section{Amortised Bayesian inference of fBM}

\subsection{fBM : a tractable random walk model}

\label{ABI}
In the context of parameter estimation, Bayesian inference 
uses Bayes' theorem to compute the posterior probability distribution of the parameters $\thetavec$ given recorded data $\Rt$ (here a trajectory) and a probabilistic model of these data,
 \begin{equation}
 \label{bayes}
     p\pare{\thetavec|\Rt}=\frac{p\pare{\Rt|\thetavec}p\pare{\thetavec}}{ p\pare{\Rt}} .
 \end{equation}
Equation~\eqref{bayes} relates the posterior distribution, $p\pare{\thetavec|\Rt}$
to the likelihood $p\pare{\Rt|\thetavec}$, the prior $p\pare{\thetavec}$ and the evidence $p\pare{\Rt}$. 
Here, we only consider one single model, i.e., the fBm, and thus do not explicitly refer to it. The principle of amortised inference~\cite{Cranmer2020} is to split the estimation of the posterior $p\pare{\thetavec|\Rt}$ into two independent steps.  The first is computationally costly and involves learning an approximate posterior density $\hat{p}\pare{\thetavec|\Rt}$ from numerically generated data. Then, the second step consists in running the pre-trained approximate system on the experimental data to infer the posterior density, assuming that they are similar to the training data.

A tractable likelihood can be computed for fBm: 
considering a trajectory $\Rt = (\r_{0}, \r_2, \ldots, \r_{N})$  to be a $1$-dimensional time-series of positions $\r_i$ recorded at equidistant points in time $t_{i} \in \lbrace 0, \Delta t, 2\Delta t, \ldots, N\Delta t \rbrace$, 
the likelihood of a trajectory reads
\begin{equation}
\label{proba}
    p\pare{\Rt|\thetavec}=\frac{1}{\pare{2\pi}^{N/2}\sqrt{\det \Sigma(\thetavec)}}\exp\pare{-\frac{1}{2} \pare{\Delta \bv{r}}^\transpose \Sigma(\thetavec)^{-1}\Delta \bv{r} } ,
\end{equation}
where  $\Delta \r = \left( \Delta \r_{1}, ..,\Delta \r_{N}  \right)^\transpose$, with $\Delta \r_{i} =  \r_{i} - \r_{i-1}$ the individual displacements. Then, $\thetavec = (K_\alpha, \alpha)$ are the fBm's parameters to infer, and $\Sigma$ is the displacements' covariance matrix whose coefficients are given by
\begin{equation}
  [\Sigma(\thetavec)]_{ij} = K_\alpha\Delta t^\alpha \left(|i-j+1|^{\alpha} + |i-j-1|^{\alpha} - 2 |i-j|^{\alpha} \right).
  \label{eq:covariance-fBm}
\end{equation}

\subsection{Amortised inference}

We choose to rely on a likelihood-free approach to amortise our inference procedure. This may seem a counter-intuitive choice for the precise case of fBm because the likelihood is analytically tractable -- other inference methods leverage this specificity, see for example \cite{thapa2022bayesian} -- but our method has the advantage of relying solely on computations of linear complexity. 
Furthermore, the approach is also directly portable to more complex problems for which a tractable likelihood may not be available or may be computationally too costly. 
Indeed, likelihood-free inference is a method of choice to address such problems. As more and more complex models are encountered in numerous fields of science, the field of simulation-based inference~\cite{Cranmer2020} is growing very rapidly to address the associated challenging inverse problems.
The shift towards amortisation of the likelihood is notably driven by new tools and conceptual approaches derived from machine learning ~\cite{pmlr-v89-papamakarios19a, Alsing2019}. Here, applying the method to a problem with a tractable likelihood allows us to compare its performance to the optimal one, derived using the Cram\'er-Rao bound.

The architecture of the amortised variational inference scheme, allowing to estimate the posterior distribution, is shown in Figure~\ref{architecture-training}. It is based on the recently introduced Bayes Flow (BF)~\cite{Radev2020} procedure. In this framework, a first neural network, the "summary network" (working as an encoder), computes a vector of summary statistics from observations. Here, the summary network is a GNN (Fig.~\ref{architecture-training}A) taking as input a trajectory and outputting a summary statistics vector $\h$ whose dimension is independent of the trajectory length. This vector then parameterises an invertible transformation between an easily sampled distribution (multivariate normal) and the posterior distribution of the parameters (Fig.~\ref{architecture-training}B). The full procedure generates a posterior distribution of the parameters. Such flow-based approaches, derived from normalising flows~\cite{Kobyzev2020}, have the advantage of providing an estimation of the posterior without requiring extensive sampling. The whole neural network is trained on numerically generated data and can then be used for inference. In the two following subsections, we first present the GNN (Section~\ref{GNN})
and then the parameterisable invertible network (Section~\ref{invertible_net}). The parameters of both these parts of the neural network are set during an upfront training phase, detailed below, during which fBM trajectories having the same parameters as those on which we seek to later perform inference are presented to the network.

\subsection{Graph neural network for learning summary statistics}
\label{GNN}

GNNs have been introduced to model and analyse graphs, meshes and point clouds~\cite{1903.02428,1609.02907,Qi2017PointNetDH}. They are well suited to capture geometric properties from point clouds and other datasets of variable size~\cite{Charles2017,Qi2017PointNetDH}, 
they can keep a sparse architecture while encoding long temporal correlations~\cite{Azevedo2020,2101.12465} 
and they exhibit good performance with a limited number of parameters compared to other modern architectures.
\change{In the following, we present the specific GNN architecture used for the summary network in this work. It is an updated version of the GRATIN architecture (as "Graphs on Trajectories for Inference") which we introduced in \cite{Verdier_2021}. We summarise in \ref{sm:differences_PRA} the differences between this version of the GRATIN architecture and that of the encoder of \cite{Verdier_2021}. We refer to this previous work and references therein the reader interested by more thorough explanations about graph convolutions, which are the core operations performed by GNNs. Throughout this paper, we use the GRATIN acronym to refer both to the architecture of the summary statistics network and to the inference scheme as a whole.}

As indicated by their name, GNNs process graphs, and the first step of our inference is thus to build a graph from a trajectory. 
To do so, we represent each trajectory $\Rt = (\r_{0},\r_{1}, \r_2, \ldots, \r_{N})$ by a directed graph $G = (V,E,\Xt,\Yt)$. Here $V = \{1, 2, \ldots, N\}$ is the set of nodes, each corresponding to a recorded position of the observed walker. 
$E \subseteq \{(i,j) | (i,j) \in V^2\}$ is the set of edges connecting pairs of nodes. 

A vector of features, $\xVec_i^{(0)}$ (of size $n_x$), is first attached to each node, encapsulating information relating to the $i$-th position of the trajectory and to the displacements that led to it. 
\change{
There are six such node features (i.e. $n_x=6$): the normalised time $i/N$, the maximal jump size since the beginning of the trajectory $\Rt$ (normalised by $s(\Rt)$, the standard deviation of jump sizes), the distance to origin and the maximal distance to origin up to point $i$, both normalised by $s(\Rt) \sqrt{i}$, which is proportional to the square root of the expected square displacement of a Brownian walker over this time span. These are complemented with three features indicative of the moments of the distribution of step sizes observed up to node $i$, detailed in supplementary \ref{sm:features}.
Thus, $\Xt^{(0)} = (\xVec_1^{(0)}, \xVec_2^{(0)}, \ldots, \xVec_N^{(0)})$ is the $(N,n_x)$ matrix of initial node feature vectors.

Similarly, $\Y = (\y_1^{(0)}, \y_2^{(0)}, \ldots, \y_{|E|}^{(0)})$ is a matrix of  edge features, $\y_{e}^{(0)}$, each associated to an edge $e$ in $E$.
The features vector of a given edge $e  = (i,j)$, $\y_{e}^{(0)}$, of size $n_y = 6$, encapsulates information about the trajectory's course between the two nodes $i$ and $j$ it connects. More precisely, edge features are the time difference $j-i$, the distance between its extremities normalised by $s(\Rt)\sqrt{j-i}$, the correlation of jumps normalised by $s(\Rt)^2$, as well as three other features related to the moments of the distribution of step sizes observed between nodes $i$ and $j$.
}

The edges in G are chosen such that incoming edges of each node originate only from nodes in the past (i.e., respecting causality): node $i$ receives connections from nodes \change{$i - \lfloor i^{k/(d-1)} \rfloor$, for $k \in \lbrace{0,1,\dots,d-1} \rbrace$ where $\lfloor \cdot \rfloor$ denotes the integer part and $d$ is the maximum incoming degree parameter: each node receives a maximum of $d$ edges, and they are thus less than $dN$ edges in the graph (in the following, we chose $d=20$). 
This scale-free wiring of nodes allows the network to capture long range correlations, and provides a good control on the way the networks structures the data. The graph structure furthermore allows to summarize the geometry of the trajectory without specifying coordinates, by resorting to distances (carried by the edges). 
Thus, the graph representation of trajectories (with nodes and edge features) taken as input by the GRATIN summary network, is rotation-invariant. This is an important advantage given that the inference problem indeed has a such invariance -- respecting symmetries of the data does in general help neural networks to converge \cite{gerken2021geometric, lecun2015deep}}

While the training is specific to the dimension of the random walk, the GNN architecture can keep the same features initialisation and number of parameters.

A key point about the graph construction procedure \change{(wiring and features initialisation)} is that \change{its algorithmic complexity is linear in the number of nodes. This is required by the objective of providing an estimator of linear complexity}.

Following the graph initialisation step, the summary network performs several graph convolution operations~\cite{1606.09375,1806.08804,1801.07829}.
It then passes the learnt node feature vectors as inputs to a pooling layer that aggregates features across all nodes of a trajectory graph into a fixed-length vector. 
The vector is finally passed through a multi-layer perceptron to obtain the summary statistics vector $\h = g_\psi(\Rt)$, where $\psi$ denotes the neural network coefficients, set during training. 
\change{In section \ref{perf}, we show inference results done on trajectories longer than those seen by the networks during their training. While the architecture is capable to handle trajectories of any size, it will not produce reliable results on trajectories which have a such major difference with those seen during the training phase. Hence, we segment long trajectories in segments of length $N \leq 1~000$ and average the vectors of summary statistics computed from these segments to obtain a summary statistics vector describing the complete trajectory.}

\begin{figure*}[!tp]
    \centering
    \includegraphics[width=.74\textwidth]{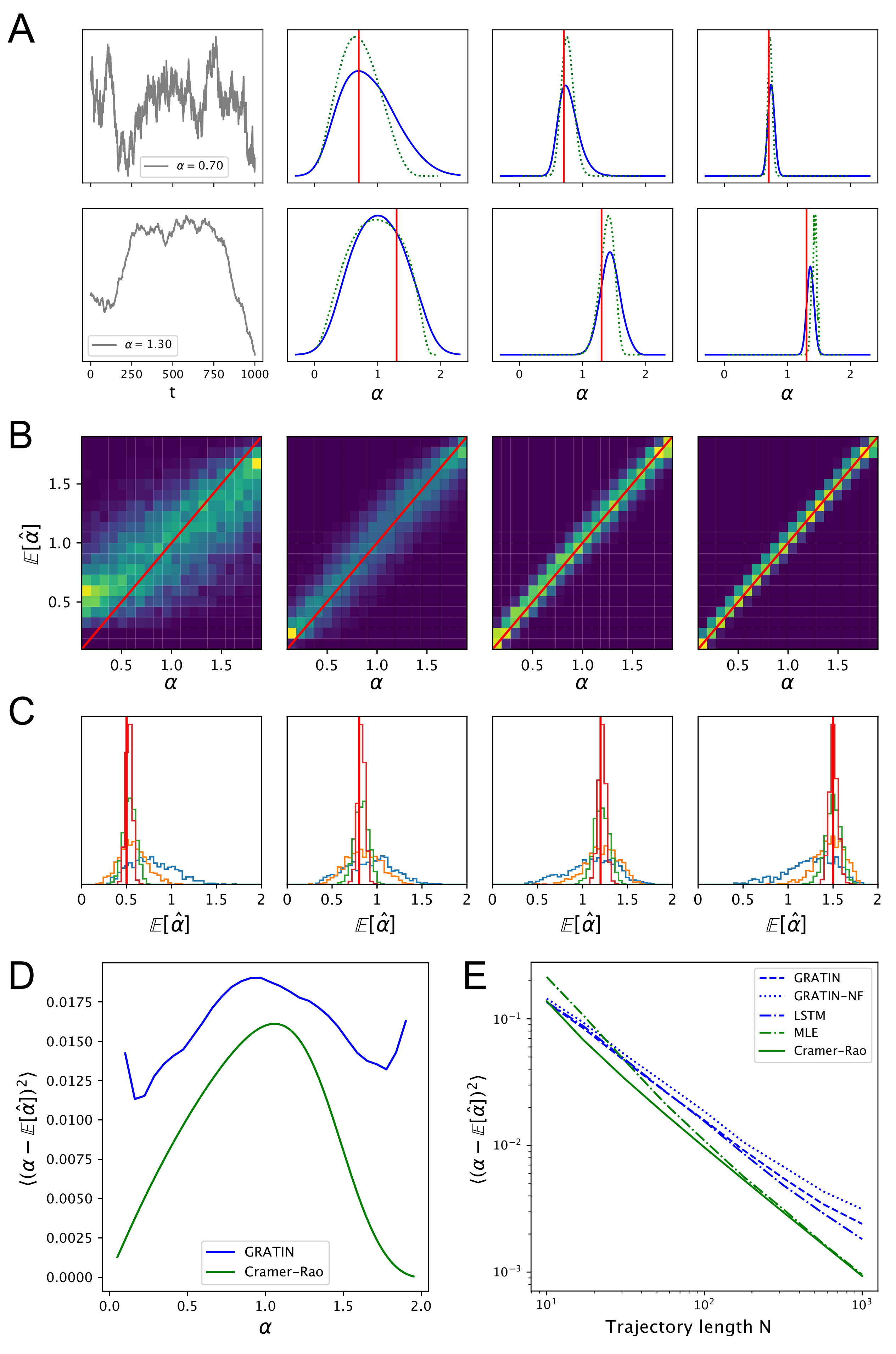}
    \caption{{\bf Model performance.}
    A: Evolution of the posterior density of $\alpha$ inferred by the model (plain blue lines) versus the true posterior (dotted green lines) from two example trajectories with $\alpha = 0.7$ (top) and $\alpha = 1.3$ (bottom), respectively. 
    The length of the portion of the trajectory used for inference is increased ten-fold between each panel, i.e., from left to right: $N$ = 10, 100, 1~000.
    B: Density of  $(\alpha, \mathbb{E}\left[ \hat{\alpha} \right])$ for 10~000 trajectories of different lengths. From left to right, $N =$ 10,50, 250, 1~000.
    C: histograms of $\mathbb{E}\left[ \hat{\alpha} \right]$ for several values of $\alpha$ (from left to right, $\alpha =$ 0.5, 0.8,1.2,1.5, marked by the red vertical lines), with trajectories of different lengths ($N =$ 10 (blue), 50 (orange), 250 (green), 1~000 (red)). Histograms are representative of $10~000$ trajectories each.
    D: Comparison of the MSE of our estimator of $\alpha$ with the optimal variance given by the Cramer-Rao bound, over the range of values taken by alpha. For each of 30 values of alpha, the MSE is an average over 10~000 trajectories.)
    E: Blue curves: evolution, with the length $N$ of trajectories, of the mean square error (MSE) of the mean posterior $\mathbb{E}\left[ \hat{\alpha} \right]$ obtained with several architectures of summary networks. for each value of $N$, 100~000 trajectories are used to compute the averages. 
    Green curves: Cram\'er-Rao bound for an unbiased estimator of $\alpha$, and MSE of the $\alpha$ of maximum-likelihood (on an evenly spaced grid of 200 values between 0.1 and 1.9).}
    \label{fig:alpha}
\end{figure*}

\subsection{Invertible network for generating a variational posterior density}
\label{invertible_net}

The Bayes Flow approach relies on an invertible transformation, $f_\phi(\cdot; \h)$, between the parameter space (in $\mathbb{R}^\nparams$, with $\nparams \geq 2$) 
and the prior space (in $\mathbb{R}^\nparams$), on which a $\nparams$-dimensional standard Gaussian density is assumed. \change{This allows the model to output a full posterior distribution over the space of parameters, while most machine learning-based techniques aimed at analyzing random walks usually restrict to point estimates.}
The transformation $f_\phi(\cdot; \h)$ is parameterised by a conditional invertible neural network (cINN)~\cite{ardizzone_analyzing_2019} made of a succession of affine coupling blocks~\cite{dinh_density_2017} (multiple blocks sequentially applied) and maps $\thetavec$ to the prior conditioned on $\h$, the summary statistics of the trajectory.

By design, these blocks can be inverted and the determinant of the Jacobian matrix $\mathbf{J}_{f_{\phi}}$ of the transformation retrieved from the forward pass. During training we seek to approximate the true posterior $p\pare{\theta|\bf{R}}$ by the learnt posterior $p_{\phi}\pare{\theta|\bf{R}} = \exp \left( - \frac{ \lVert f_\phi(\thetavec; \h) \rVert^2_2}{2} \right)$. The loss function is the  
Kullback-Leibler divergence between $p\pare{\theta|\bf{R}}$ and $p_{\phi}\pare{\theta|\bf{R}}$ which reads as 
\begin{equation}
    \label{eq:objective}
    \mathcal{L}(\Rt) = \frac{1}{2}\lVert f_\phi \left(\thetavec; \h \right ) \rVert^2_2 - \log |\det \mathbf{J}_{f_{\phi}}| ,
\end{equation}
where $\h = g_\psi(\Rt)$ and $\lVert \cdot \rVert_2$ denotes the Euclidean norm. Sampling the posterior distribution first requires computing $\h$ from the trajectory $\Rt$, and then generating the required number of sample as $\thetavec = f^{-1}_\phi(\z; \h)$  with
$\z$ drawn from a standard $D$-dimensional Gaussian distribution. 

\change{Details about the implementation are presented in Supplementary Material 1., and the code is available online : \href{https://gitlab.pasteur.fr/hverdier/gratin-fbm}{https://gitlab.pasteur.fr/hverdier/gratin-fbm}}



\section{Estimation of the anomalous exponent}
\label{perf}

We evaluate the performance of our variational inference procedure on numerically generated trajectories.
Estimating the anomalous exponent $\alpha$ is the most challenging part of the inference, and we thus focus on this here, although our approach infers a joint posterior density for $\thetavec = (K_\alpha, \alpha)$. \change{Trajectories used for training and evaluation are generated using the Davies-Harte method \cite{davies1987tests} implemented in the Python \texttt{fbm} package with $\alpha \sim \mathcal{U}(0.1,1.9)$ and $\log_{10}(K_\alpha) \sim \mathcal{U}(-2,2)$}.

Figure~\ref{fig:alpha}A shows the inferred posteriors of $\alpha$ on portions of increasing length of two example trajectories. 
The amortised posterior is consistent with the exact one, and both become increasingly peaked around the true value of $\alpha$ as the length of the trajectory increases.
The inferred posterior distributions do not exhibit broad tails or divergences, and are proper distributions, i.e., they are normalisable.

Using the likelihood shown in ~\ref{proba} we computed the Cram\'er-Rao bound of the inference problem, which gives a lower bound of the variance achievable by an unbiased estimator of the parameters $\thetavec$ (see Supplementary Material \ref{sm:cramer_rao}). We computed $b_N(\alpha)$ the lower bound of the variance on $\alpha$ for estimators processing fBM realisations of length $N$ with an anomalous diffusion exponent $\alpha$. 
Besides, reducing for each trajectory the inferred posterior distribution to its mean, $\mathbb{E}\left[ \hat{\alpha} \right]$, we compute the mean square error (MSE) $\langle \left( \mathbb{E}\left[ \hat{\alpha} \right] - \alpha \right)^2 \rangle$ of the inference.
On Fig. \ref{fig:alpha}D, we compare this quantity (for $N=100$) to the MSE of our amortized inference based on the GRATIN summary network. We observe that the optimal variance critically depends on $\alpha$ and that our inference is close from optimal on trajectories with $\alpha \approx 1$.

\begin{figure*}[!tp]
    \centering
    \includegraphics[width=.89\textwidth]{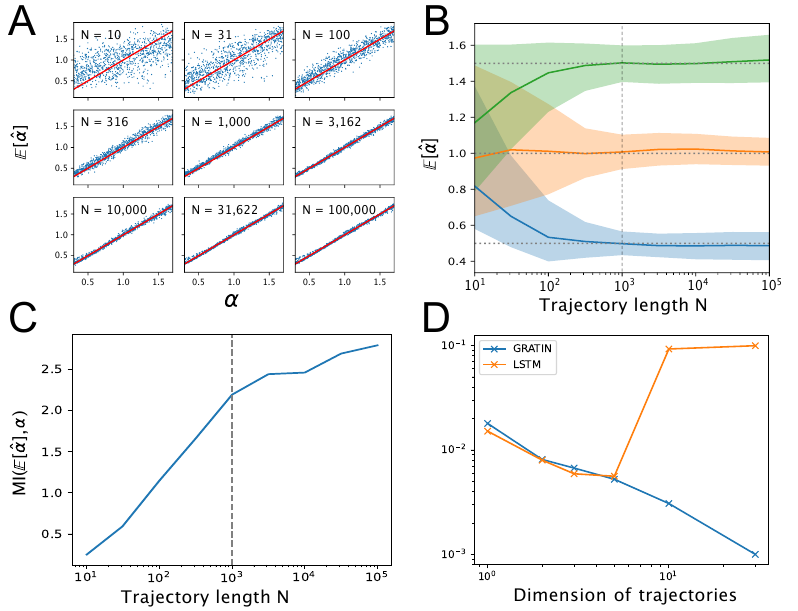}
    \caption{
    A: Comparison of mean inferred values and true values of $\alpha$, for trajectory lengths spanning four orders of magnitude. The red line is the diagonal $\alpha = \mathbb{E}\left[ \hat{\alpha} \right]$.
    B: Convergence towards the true value of $\alpha$ as function of trajectory length. The model was trained on trajectories of length $10 \leq N \leq 1~000$; longer trajectories were cut in segments of length $\leq 1~000$ and their summary vectors averaged to obtain a single posterior for the whole trajectory. Shaded areas represent the 5\% \& 95\% quantiles.
    C: Mutual information between $\alpha$ and $\mathbb{E}\left[ \hat{\alpha} \right]$. 
    D: Performance of models based on GRATIN and LSTM encoders as a function of the trajectory dimension ($D=1, 2, 3, 5, 10, 30$). 
    }
    \label{fig:MI_highD}
\end{figure*}

In Fig. \ref{fig:alpha}E, we show the mean MSE of our inference on trajectories with lengths varying across two orders of magnitudes, with $\alpha$ uniformly sampled in the interval $(0.1,1.9)$. We observe that the MSE follows a power-law decrease and is close to the mean optimal variance (obtained by averaging, for each $N$ the values of $b_N(\alpha)$ over the range of values of $\alpha$). 
The amortised inference is suboptimal (as expected from any variational inference), but its variance shows a fast decreasing trend similar to the Cram\'er-Rao bound, i.e., close to $\propto 1/N$.
\change{We compared several architectures of summary networks : GRATIN, which is described above, GRATIN-NF (as "no features"), which is a version of GRATIN with a minimal set of features (see supplementary material \ref{sm:features}) and an architecture based on 1D convolutions and LSTM layers ("Long Short Term Memory", \cite{lstm}), inspired from \cite{argun2021classification} and \cite{garibo2021efficient}, who proposed similar architectures to study anomalous diffusion (see details in supplementary material \ref{sm:lstm}). We observe that features do improve the quality of the inference performed by GRATIN. The LSTM summary network provides slightly more precise predictions than the GRATIN one, but this is at the cost of a lesser interpretability of the inner computation performed by the network and a greater number of parameters. 

An advantage of a BayesFlow-based framework is that, by averaging summary statistics vectors of segments of long trajectories, the network is able to extract information about longer trajectories than those seen during training. We show in figure \ref{fig:MI_highD}A that the spread of mean inferred predictions keeps shrinking as $N$ increases. We quantify this in figure \ref{fig:MI_highD}B, where the evolution of the mutual information (MI) between the true and the mean inferred $\alpha$ is shown as a function of trajectory length. To estimate MI, we used the procedure described in \cite{kraskov2004estimating} implemented in \cite{scikit-learn}. Nevertheless, as segments of a long trajectories are correlated, so are their summary statistics vectors. Hence, their average does not necessarily lay in the same region of the latent space as representations of segments of same size taken from individual trajectories with the same value of $\alpha$. This results in a slight systematic bias of the prediction, which we corrected using a simple polynomial.

\change{
The rotation invariance which the GNN enforces is especially relevant on trajectories of higher dimension. To confirm the interest of a such constraint, we have trained inference models based on GRATIN and LSTM summary networks on trajectories of increasing dimension (and fixed length $N=100$) and observed that models based on GRATIN yielded better results than LSTMs for trajectories of dimension greater than three, as shown in figure \ref{fig:MI_highD}D}. The LSTM architecture seems to encounter convergence issue when processing trajectories of high dimensions. We kept the same hyperparameters of training throughout all experiments, i.e. a batch size of 2~048 and a learning rate of 0.01, which is reduced by a factor 0.2 when the loss plateaus for 15 epochs. Training was stopped after 30 epochs of 100,000 trajectories. While a thorough hyperparameter optimization could reveal the configuration which enables the LSTM architecture to properly converge, this highlights the robustness of the GNN architecture. Note that interest in GNN is based on physical interpretability of the latent space, linear complexity with trajectory length and the possibility to implement physically relevant symmetries within the Graph structure. There is little doubt that numerous machine learning architecture may be found to be more efficient in specific point perdiction processes. }

The learnt summary statistics $\h$, used to compute the posterior distribution, constitutes a low-dimensional representation of a trajectory . The structure of the manifold on which summary statistics vector lie is revealing of the way the encoder represents information extracted from the trajectories. An assumption in representation learning~\cite{bengio_representation_2014} is that interpretable representation lead to better generalisation. We projected $\h$ onto a 2D plane using UMAP~\cite{mcinnes2018umap} (a non-linear dimensionality reduction algorithm) and mapped $\alpha$ on it (see Supplementary Fig. S1). We see that the latent space is organised according the value of $\alpha$, a good indication that the learning process properly captured the underlying physical properties. 
We tested the robustness of the inference procedure when applied to trajectories corrupted by positional noise. We show in Supplementary Fig. S2 the evolution of the MSE of the amortised inference of $\alpha$ and compare it with the corresponding Cram\'er-Rao bound. The precision of the amortised inference procedure closely follows the lower bound set by the Cram\'er-Rao inequality. This was obtained by training models specifically on trajectories corrupted with increasing amounts of noise.


\change{An important attribute of our approach is that, passed the amortisation (training) step, the computational complexity of inferring the parameters of a novel trajectory is linear in terms of the length of this trajectory.}
To show this, we subdivide the inference procedure into three steps: (i) initial feature evaluation, (ii) forward pass through graph convolutions and pooling, and (iii) operations on summary statistics to generate the posterior. 
(i) The initial evaluation of node and edge features requires $O(N+|E|)$ time and memory, where $N$ is the number of nodes (for a trajectory of $N+1$ points) and $|E|$ is the number of edges. Here $|E| \propto N$ by design (the in-degree of nodes is bounded), so this step has $O(N)$ complexity. 
(ii) The forward pass through the graph convolutions and the following pooling of node features requires $O(|E|)$ operations and memory slots, and hence this step also has $O(N)$ complexity. 
(iii) The latent space is of fixed dimensions, and hence all operations after the pooling layer have $O(1)$ complexity.
The global complexity of the amortised architecture is thus linear with respect to the number of points in the trajectory.

In comparison,  calculating the exact likelihood [Eq.~\eqref{proba}] requires evaluating the determinant $\mathrm{det}\Sigma(\thetavec)$ and the quadratic form $(\Delta\bv{r})^\transpose \Sigma(\thetavec)^{-1} \Delta\bv{r}$, which can be done in $O(N^2)$ time~\cite{Monahan_2011}.
This makes exact inference prohibitively expensive for very long trajectories, where our amortised inference scheme may instead be used (Fig.~\ref{fig:alpha}C).
Note furthermore that for many models the exact likelihood cannot be calculated at all, in which case approximate inference is the only route possible.


\section{Estimation of a finite decorrelation time}
\label{sec:decorrelation}

When considering fBm as a model of biomolecule random walks, we have to keep in mind that many physical environments  might exhibit a finite decorrelation time $\tau_c$ possibly stemming from motion occurring outside a polymer-dominated environment \cite{weber_bacterial_2010} or from changes of conformations of the bimolecule altering the nature of its interactions. The characteristic time bears information on the local environment's physical properties, and it might be spatially dependent or specific to interactions with local partners. 
In practice, inferring $\tau_c$ from individual trajectories is challenging. Autocorrelation-based approaches for example give incomplete results on individual trajectories as the limited number of points prevents proper averaging~\cite{reverey_superdiffusion_2015,crispin_handbook_2009}. 
\change{The power spectrum density (PSD) is a relevant quantity as well, but can only be estimated with sufficient precision for trajectories longer than those considered here by several orders of magnitude. Besides, \cite{krapf2019spectral} show that PSD curves computed on an ensemble of superdiffusive fBM trajectories cannot be simply averaged, hence the interest of inferring $\tau_c$ directly from single trajectories.}

We adapted the amortised inference procedure presented above to infer $\pare{\alpha,\tau_c}$ instead of  $(K_\alpha,\alpha)$. 
We left out $K_\alpha$ here since it is simply a scale factor and can be removed by rescaling the trajectories.
We used the same node and edge features as above, ensuring the procedure's linear computational complexity. 
A finite decorrelation time was modeled by multiplying the autocovariance of the fBm 
by an exponential factor, $\text{min}(1,e^{\tau_c-\tau})$, where $\tau$ is the time difference. 
Examples of the autocorrelation function for several values of $\alpha$ and $\tau_c$ are given in Supplementary Fig. S3. 
The modified covariance matrix of the displacements thus reads
\begin{eqnarray*}
  [\Sigma(\alpha,\tau_c)]_{i,j} &&=
  \text{min}\left(1,e^{\tau_c-|i-j|\Delta t}\right) \times \\ &&\left(|i-j+1|^{\alpha} + |i-j-1|^{\alpha} - 2 |i-j|^{\alpha} \right) ,
\end{eqnarray*}
where we have ignored the scale factor $K_\alpha$.  As there is no simple means to quantitatively relate the length of a trajectory to the difficulty of inferring its finite correlation time, we performed this inference on trajectories of fixed length $N=$ 1~000, with $\tau_c$ integer-valued and ranging from 5 to 50. We chose this value of $N$ firstly because the length of the trajectory must exceed the correlation time by at least an order of magnitude in order for some information to be retrievable and secondly because the simulation of trajectories has a quadratic complexity in $N$, deterring one to include too long trajectories in the simulation-greedy training phase. 
\change{We chose as priors $\log \left(\tau_c\right) \sim \mathcal{U}\left(\log(5),\log(50)\right)$ and $\alpha \sim \mathcal{U}\left(0.4,1.6)\right)$. We restricted the range of $\alpha$ compared to the previous sections because the modified covariance matrices were not invertible for extreme values of $\alpha$. We simulated trajectories by sampling the steps from the multivariate Gaussian distribution $\mathcal{N}(\mathbf{0},\Sigma(\alpha,\tau_c))$ using a Cholevsky decomposition : $\Delta \mathbf{r} = L \mathbf{u}$, where $L$ is the Cholevsky decomposition of $\Sigma(\alpha,\tau_c)$ and $\mathbf{u}$ is a Gaussian vector drawn from the multivariate normal}. 

\begin{figure*}[!tp]
    \centering
    \includegraphics[width=.95\textwidth]{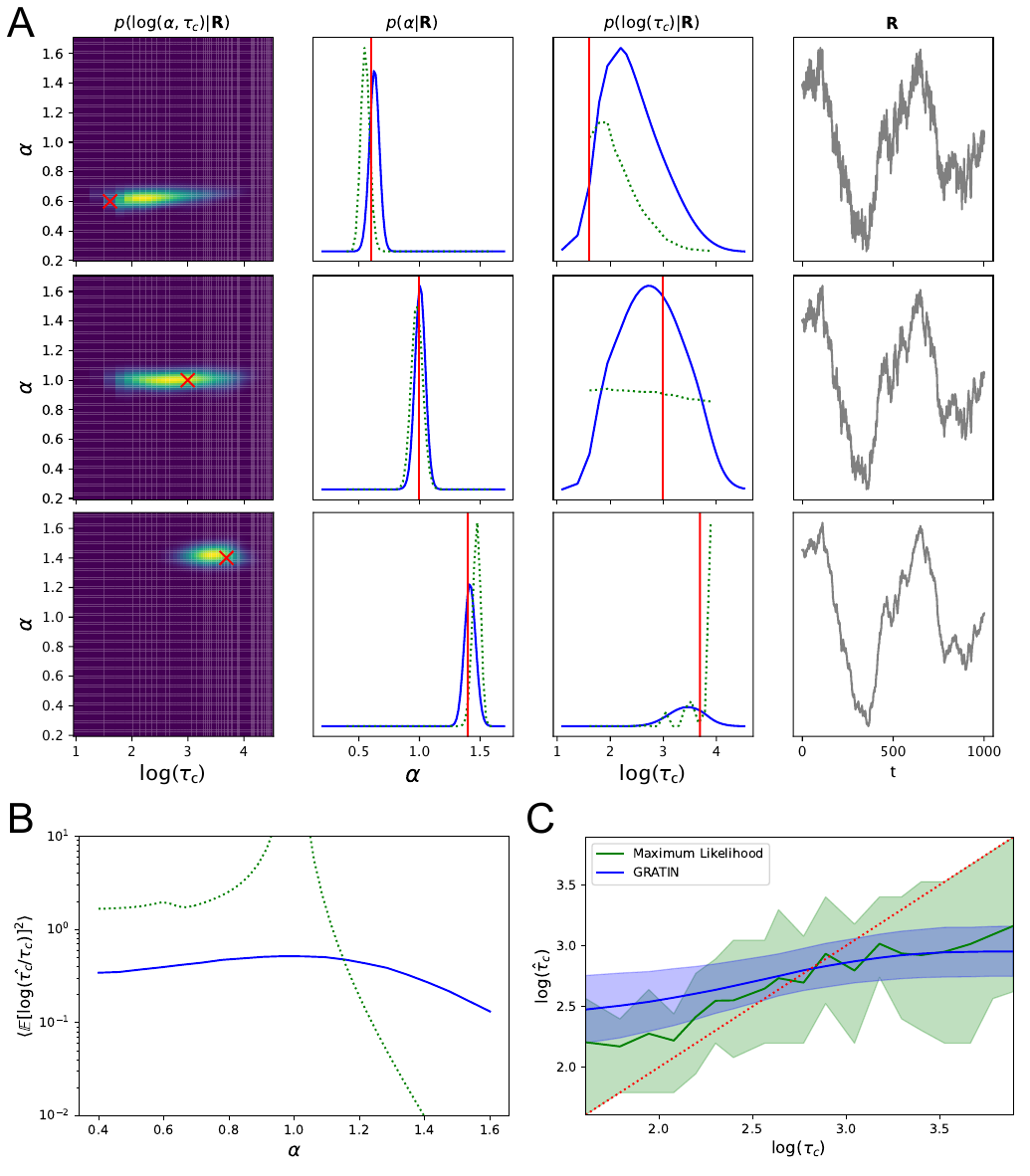}
    \caption{{\bf Performance of the anomalous exponent and decorrelation time estimation}
    A: left column : joint posteriors of $\alpha$ and $\log_{10}(\tau_c)$, middle columns : marginal inferred posteriors (plain lines) and true ones (dashed lines), right column: corresponding trajectories. True values of the parameters are indicated in red. From top to bottom, $(\alpha, \tau_c)$ equals $(0.6,5)$, $(1,20)$, $(1.4,40)$. 
    B: Variance and MSE of our inference of $\log(\tau_c)$ plotted as a function of $\alpha$, compared with the Cram\'er-Rao bound for an unbiased estimator. For each of 20 values of $\alpha$, 600 trajectories with $\log(\tau_c)$ sampled uniformly between $\log(5)$ and $\log(50)$ were used to compute the mean square error on the inference of $\log(\tau_c)$.
    C: Comparison of the values of $\log(\tau_c)$ inferred by our method (in blue) and by a maximum-likelihood (ML) estimator (in green). The thick line represents the mean across all trajectories, while the filled regions correspond to the first and last quartiles. The anomalous diffusion exponent $\alpha$ is drawn from a uniform distribution between 0.4 and 1.6. We considered, for each value of $\tau_c$, 100 trajectories  for the ML estimation (computationally intensive) and 6,000 for our amortized inference. The most plausible values of $\tau_c$ were chosen on a grid of 20 values of $\alpha$ and 20 values of $\tau_c$.}
    \label{fig:alpha_tau}
\end{figure*}

\change{
Examples of inferred joint posteriors of $(\alpha, \log(\tau_c))$ are shown in Fig. \ref{fig:alpha_tau}A. The precision of the inference of $\alpha$ remains high despite the limited correlations, while the order of magnitude of $\tau_c$ is often retrieved (except around $\alpha \approx 1$). The example trajectories shown on the rightmost column give a good intuition of the effect of the finite correlation time as they were generated with the same random vector $\mathbf{u}$ (only the autocovariance matrix varied): all three trajectories have the same global shape, only fine-grained dynamics reveal the difference of anomalous exponent.}

We compared our estimator with the maximum-likelihood one, which consists in choosing the value of ($\alpha$,$\tau_c$) that maximises the likelihood of the observed trajectory. In practice, we computed the log-likelihood on a grid of values of $\alpha$ and $\tau_c$ to find the values yielding the maximal likelihood, with $\alpha$ taking 30 regularly spaced values between 0.4 and 1.6, and $\tau_c$ taking 25 geometrically-spaced values in its range. As shown in Figure~\ref{fig:alpha_tau}C, our amortised inference yields a slightly more biased estimate than the maximum-likelihood estimator (when taking the mean of the posterior distribution) but has a smaller variance. 
When $\alpha = 1$, successive increments are completely independent of each other and there is thus no information to retrieve regarding $\tau_c$. This is observable on the lower panel of Figure \ref{fig:alpha_tau},  both by looking at the Cram\'er-Rao bound, which diverges, and at the variance of our estimator, which is maximal at $\alpha = 1$.


\section{Discussion}
\change{
Simulation-based inference coupled with machine learning are a promising avenue to address challenging inverse problems. When applied to intractable systems (i.e., systems for which no closed-form expression of the likelihood is known), this combination allows splitting the inference task into two steps.  In the first phase, computationally intensive, simulations produce artificial data on which a neural network is trained to approximate the posterior distribution of the parameters using a variational objective. In the second step, which is computationally fast, inference is performed on experimental data by a direct forward pass through the trained network, yielding the estimated posterior distribution. 
The procedure is statistically efficient if the numerical data match the properties of experimental one and if the variational inference is able to capture the complex relations that might exist between the variables to be inferred. }

There are two main challenges associated with amortised approaches. 
First, training variational inference often consists in minimising a Kullback-Leibler divergence between the approximate distribution and the real (unknown) one ~\cite{bishop_pattern_2011}. Optimising such a non-convex function is challenging and is not generally  guaranteed to converge towards a global optimum.
The second challenge is linked to interpretability. Both the models used to learn the summary statistics and the variational posterior distribution are generally intractable.
There is thus limited insurance that the process does not misbehave, especially when applied to real experimental data. 
Evaluation of the exact posterior distribution using sampling, such as in approximate Bayesian computation, may however lead to similar problems due to the difficulty of properly sampling complex likelihood landscapes. \change{Discrepancies between the data on which the inference is to be run and the simulated data used for training, likely to harm the inference's relevance, could be detected using methods such as the one we introduce in \cite{verdier2022maximum}, based on statistical tests between summary vectors.}

We here used fBm to quantify the performance of our amortised inference approach. 
We chose to focus on fBm both due to its paradigmatic status as an anomalous random walk model and because it has a tractable likelihood, allowing us to compare our amortised method to exact likelihood-based inference and to the Cram\'er-Rao bound on estimator precision. 
We advocate more generally for the use of exactly solvable random walk models, such as the fBm, as benchmarks to evaluate the performance of machine-learning based inference methods. 

We showed that our amortised inference can successfully be applied to infer the parameters of fBm, with a precision that is lower than the Cram\'er-Rao bound but which increases with a scaling that is similar to it. 
Our algorithm has a linear complexity in the length of the trajectory and can be applied to trajectories of any length at inference time, even if the algorithm has not been specifically trained on trajectories of the same length.  
We furthermore showed that our amortised approach could be used to efficiently infer the parameters of a more realistic fBm-type model with a finite decorrelation time.

Our amortised inference framework can be used for any random walk model, even for models that do not have a tractable likelihood, provided that they can be simulated efficiently enough to provide a large number of trajectories for training.
In all cases, our approach retains its linear computational complexity at inference time.
For random walk models with intractable likelihoods, only empirical evaluation of the performance will in general be possible. Thus, it is not possible to make absolute statements about the statistical efficiency of the approach in these cases.   

Beyond random walks, amortised inference can more generally be instrumental in providing posterior distributions for models of complex systems with fractional noise and/or long memory. 
Recent work by the authors of BayesFlow has extended its scope of application to model comparison problems \cite{radev2021amortized}. 
Numerous challenges remain to be addressed to standardise the optimisation of the variational inference, especially in cases where some parameters are not sufficiently constrained by the data or when there are sloppy directions in the parameter space~\cite{waterfall_sloppy-model_2006}. 
Furthermore, variational inference does not necessarily lead to physically realistic parameters. Ensuring the physics-informed \cite{raissi_physics-informed_2019,both_fully_2021} nature of the inference may require imposing constraints on the network generating the summary statistics, but our results show that the network is able to learn physically meaningful features without inductive bias.
Finally, the statistical efficiency of amortised approaches will depend on the ability of numerically generated data to match experimental observations.


\vskip 0.3in
\textbf{Acknowledgments.}
 We thank Thomas Blanc, Mohamed El Beheiry, Srini Turaga, Hugues Berry, Raphael Voituriez $\&$ Bassam Hajj for helpful discussions.
This study was funded by the Institut Pasteur, \emph{L'Agence Nationale de la Recherche}~(TRamWAy, ANR-17-CE23-0016), the INCEPTION project (PIA/ANR-16-CONV-0005, OG), and the \emph{``Investissements d'avenir"} programme under the management of Agence Nationale de la Recherche, reference ANR-19-P3IA-0001 (PRAIRIE 3IA Institute).  

The funding sources had no role in study design, data collection and analysis, decision to publish, or preparation of the manuscript.

\textbf{Conflicts of interest.}
Hippolyte Verdier and Alhassan Cass\'e are Sanofi employees and may hold shares and/or stock options in the company.The other author declare to have no financial or non-financial conflicts of interest.







\clearpage
\appendix


\section*{Supplementary Material}

\subsection{amortised inference model architecture and training}
\label{sec:gnn_detail}

\subsubsection{Node and edge features}
\label{sm:features}
\change{

The features associated to nodes and edges of a trajectory's graph in the GRATIN summary network are defined using the following variables defined for each node $i \in \lbrace 1,\dots,N\rbrace$: the sum $R_i^{(k)}$ of step sizes elevated to the power $k$, up to point $i$: $\sum_{j \leq i}\lVert \Delta \r_{j}^{k} \rVert_2$. We denote as well $s$ the standard deviation of step sizes.
Then, the node features associated to a node $i$ are:
\begin{enumerate}
    \item the normalised time: $i/N$,
    \item the normalised distance to origin $\lVert \r_i \rVert_2 / s \sqrt{i} $,
    \item the normalised maximal distance to origin up to point $i$ : $\max_{k\leq i} \lVert \r_k \rVert_2 / s \sqrt{i} $,
    \item normalised cumulative powered distances covered by the walker up to point $i$, with $k \in \lbrace 1, 2, 4 \rbrace$: $\frac{N}{i}\frac{R_i^{(k)}}{R_N^{(k)}}$.
\end{enumerate}
The six features associated to an edge $(i,j)$ with $i < j$ are:
\begin{enumerate}
    \item the time difference: $j - i$,
    \item the normalised distance between edge source and target: $ \lVert \r_j - \r_i \rVert_2 / s \sqrt{j-i}$,
    \item the normalised dot product of jumps: $\Delta \r_i^\transpose \Delta \r_j / s^2$,
    \item normalised differences of the cumulative powered distances covered by the walker from point $i$ to $j$, with $k \in \lbrace 1, 2, 4 \rbrace$: $\frac{N}{j-i}\frac{R_j^{(k)} - R_i^{(k)}}{R_N^{(k)}}$.
\end{enumerate}
The number of operations required to compute features scales linearly with the trajectory length.

The GRATIN-NF version of the summary network simply uses two features per node and edge. Its node features are :
\begin{enumerate}
    \item the normalised time: $i/N$,
    \item the normalised distance to origin: $\lVert \r_i \rVert_2 / s \sqrt{i}$.
\end{enumerate}
Its edge features are :
\begin{enumerate}
    \item the normalised time difference : $(j-i)/N$,
    \item the normalised distance : $\lVert \r_i - \r_j \rVert_2 / s \sqrt{j-i}$.
\end{enumerate}
}

\subsubsection{GNN Architecture}

The architecture of the GNN used in the summary network is similar to the encoder network proposed in~\cite{Verdier_2021}, with the difference that we here additionally apply edge features. 
Node and edge features are first passed to perceptrons, which embed them in an 8-dimensional space. The network is then composed of three successive convolution layers, one relying simply on node features and the two others being conditioned by edge features. 
\change{We used as convolutions the GIN layers introduced in \cite{xu2018powerful}). They output $\xVec^{(1)}$, $\xVec^{(2)}$ and $\xVec^{(3)}$ vectors, each of 32 dimensions (equivalent to 32 convolution filters), which are concatenated} 
to form $\xVec^{(f)}$. The rows of this $(N,32 \times 3)$ matrix of nodes features are then aggregated using an attention mechanism during the pooling step, to keep just one row per graph, i.e., per trajectory. This vector is subsequently passed to a three-layer perceptron, the output of which is a 12-dimensional summary statistics vector. The dimension of the summary statistics vector is voluntarily higher than that of the parameters space so as to facilitate the neural network convergence, which is helped by over-parameterisation \cite{du2018gradient}.

\change{
\subsubsection{Differences with \citeauthor{Verdier_2021}}
\label{sm:differences_PRA}
The network introduced here performs variational inference on a well-defined class of trajectories, while the one presented in \cite{Verdier_2021} provides a point estimation of the anomalous exponent and the random walk type for a variety of random walk models, accounting for experimental noise. The main similarity between the two networks is the GNN structure of their encoders. Here is an extensive list of the differences between these two modules :
\begin{enumerate}
    \item In the present work, features are not only associated to nodes but to edges as well.
    \item We use a more expressive graph convolution layer, able to process edge features, the GIN layer introduced in \cite{xu2018powerful}.
    \item We use batch normalisation layers which are more appropriate to graphs, the InstanceNorm layer introduced in \cite{ulyanov2016instance}.
    \item The argument of the geometric series used to wire nodes is no more fixed by the length of the trajectory but instead depends on the index of the node of destination.
    
\end{enumerate}
}

\subsubsection{LSTM architecture}

The summary network based on LSTM takes as input, for each trajectory, the vector of positions. It has the following sequence of layers:
\begin{itemize}
    \item 2 1-dimensional convolutions, with 64 convolution kernels of size 8.
    \item 4 stacked bi-directional LSTM whise hidden units have 32 dimensions
    \item 1 MLP of size $(32\times2\times4,12)$. The input dimension corresponds to the product of the number of dimension of hidden units, the number of stacked layers and the number of directions. Indeed, the MLP takes as input a concatenation of the eight last hidden states of the LSTMs. The output has the dimension of the latent space.
\end{itemize}

\subsubsection{Invertible network}
The invertible network is a succession of three affine coupling blocks. These blocks, introduced in ~\cite{dinh_density_2017}, transform an input vector $\bv{u}$ into $\bv{v}$ in an invertible manner parameterised by the summary statistics vector $\bv{h}$. They do so by splitting $\bv{u}$ into two halves $\bv{u}_1$ and $\bv{u}_2$, used to compute the two halves $\bv{v}_1$ and $\bv{v}_2$ of $\bv{v}$ by consecutively performing the two following operations :
\begin{align*}
    \bv{v}_1 &= \bv{u}_1 \odot \exp \left( s_1 \left (\bv{u}_2;\bv{h} \right) \right) + t_1(\bv{u}_2;\bv{h}) \\
    \bv{v}_2 &= \bv{u}_2 \odot \exp \left( s_2 \left (\bv{v}_1;\bv{h} \right) \right) + t_2(\bv{v}_1;\bv{h})
\end{align*}
where $\odot$ denotes the element-wise multiplication (the Hadamard product) and where $s_1$, $s_2$, $t_1$ and $t_2$ are multi-layer perceptrons, which do not need to be invertible. In our case, they have five hidden layers and their activation function is an exponential linear unit. This procedure can be inverted to efficiently retrieve $\bv{u}$ from $\bv{v}$.

\subsubsection{Training the networks}

Not all parameter directions in the parameter space are equally constrained by the data.
Thus ,we split the summary statistics vector in two halves (one per parameter to infer) and pre-train summary networks to infer each parameter individually. This is motivated by the expected misbehaviour of variational optimisation for an inference whose parameters are under significantly different constraints. Hence, the good performance at inferring an easily learnt parameter (such as $\K_\alpha$) does not prevent the network from converging towards a better optimum where it infers more challenging parameters as well. 
We do this by using the output of the encoder GNN as an input to a multi-layer perceptron, and optimising the so-obtained regressor to infer a given parameter in a regression setting. The multi-layer perceptron is then discarded, and the outputs of the parameter-specific GNNs are concatenated to form the summary statistics vector used in the coupled inference. Weights of the summary networks are then frozen and only the invertible part of the network is trained.

\subsection{Exact posterior inference}

To compute exact posteriors, likelihood values were computed on grids of points in parameter space. We picked a uniform prior on $(0.4,1.6)$ for $\alpha$ and a $\log$-uniform one for $\tau_c$ and $K_{\alpha}$, which spanned 8 orders of magnitude. There was thus no coupling between parameters in the priors. The parameters used to generate trajectories during training were sampled from these same priors.

\subsection{Cram\'er-Rao bound}
\label{sm:cramer_rao}

Formally, we consider any estimator of the parameters $\thetavec$ to be a (possibly implicit) function of the recorded trajectory, $\Rt$, i.e., $\hat{\thetavec} = \Tf(\Rt)$.
We denote by $\psivec(\mathbf{\thetavec}) = E\left[\Tf(\Rt)\right]$ its expectation,
and by $\Gamma(\thetavec) = E[(\Tf(\Rt) - \psivec(\thetavec))(\Tf(\Rt) - \psivec(\thetavec))^\transpose]$ 
its covariance matrix.
Finally, $\If(\thetavec)$ is the Fischer information matrix, whose elements are given by $\If_{n,m}(\thetavec) = E\left[ \frac{\partial}{\partial \theta_n} \log p (\Rt|\thetavec) \frac{\partial}{\partial \theta_m} \log p (\Rt|\thetavec) \right]$.

The Cram\'er-Rao bound states that, for any estimator $\Tf$,
\begin{equation}
  \Gamma(\thetavec) \geq \nabla\psivec(\thetavec) \left[\If(\thetavec)\right]^{-1} \left[\nabla\psivec(\thetavec)\right]^\transpose , 
\end{equation}
where $\nabla\psivec$ is the Jacobian of $\psivec$.
In particular, this matrix inequality implies the following lower bound on the variance of any unbiased estimator of a single parameter: 
\begin{equation}
  \mathrm{Var}_{\thetavec}(T_n(\Rt)) \geq \left[\If(\thetavec)^{-1}\right]_{n,n}
\end{equation}.
This inequality is loose and the bound exaggeratedly low if parameters are not approximately equally difficult to estimate. Hence, when computing the lower bound of the variance $b_N(\alpha)$, we treated the problem of section \ref{perf} as univariate, $i.e.$ we discarded the terms related to the inference of $\K_\alpha$ in the Fischer information matrix.


\clearpage

\subsection{Supplementary figures}

\renewcommand\thefigure{S\arabic{figure}}    
\setcounter{figure}{0}

\begin{figure}[!htb]
    \centering
    \includegraphics[width=.45\textwidth]{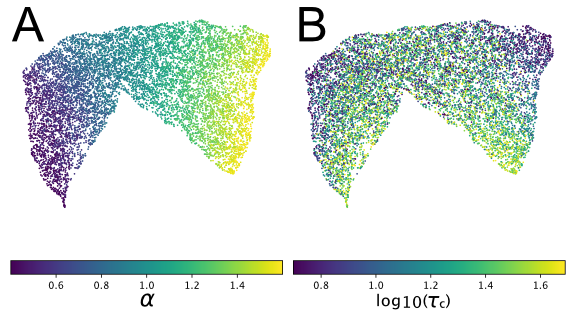}
    \caption{{\bf Latent space representations of individual trajectories.}
    2D visualisation of summary vectors (one point per trajectory), obtained by UMAP and colored according to A: their anomalous diffusion exponent $\alpha$, and B: their correlation time $\tau_c$. Trajectories are of length $N = 1,000$.}
    \label{fig:latent}
\end{figure}

\begin{figure}[!htb]
    \centering
    \includegraphics[width=.45\textwidth]{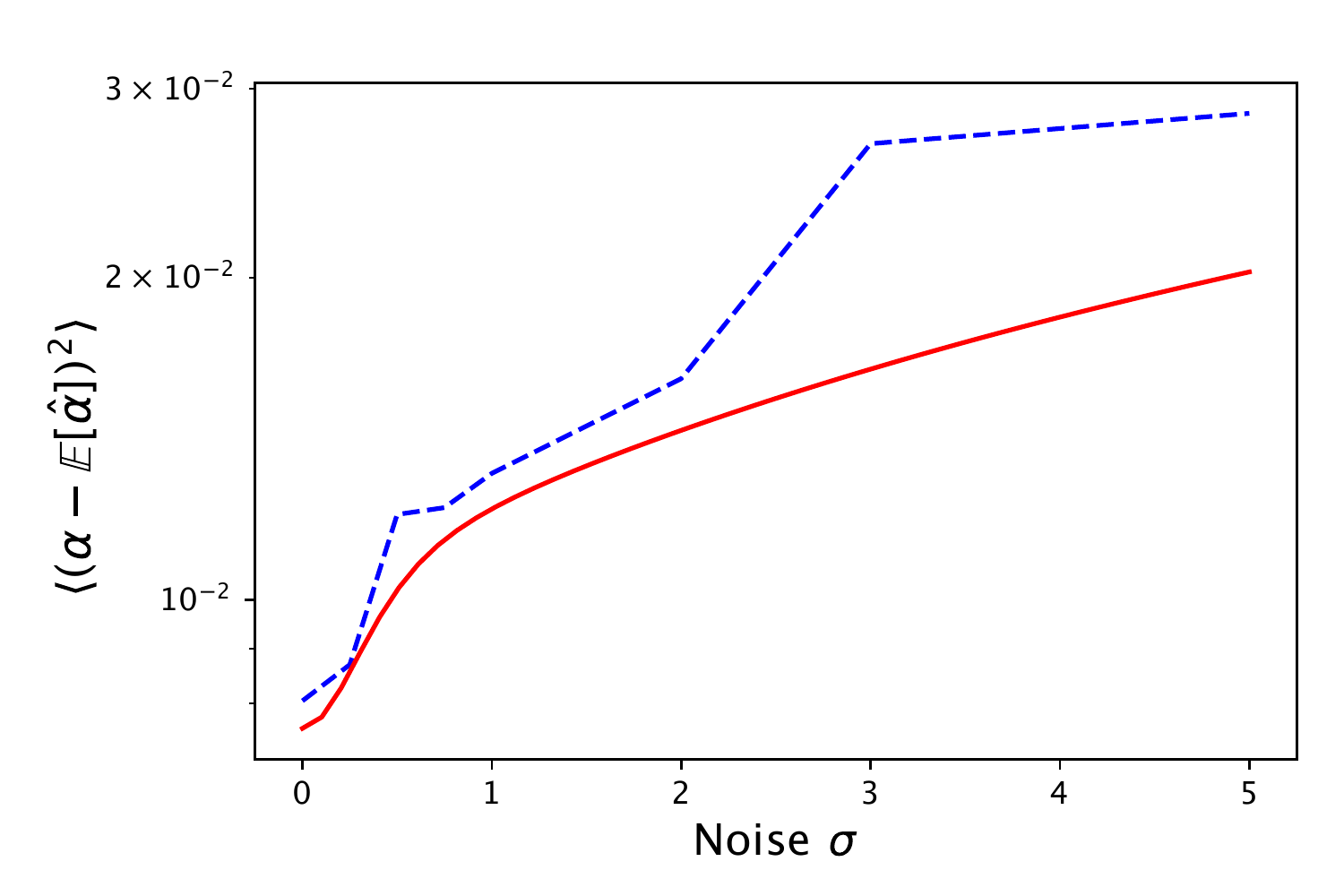}
    \caption{{\bf Robustness to noise.} 
    MSE on $\alpha$ estimated with amortised inference compared to the Cram\'er-Rao bound as a function of positioning noise $\sigma$. Trajectories are of length $N=200$ and generalised diffusivity 1. Positions are independently corrupted with Gaussian noise of variance $\sigma^2$. The diffusion coefficient was set to $1$, $\alpha$ was sampled from a uniform distribution between 0.4 and 1.6, and the performance was obtained by averaging over 10~000 trajectories for each level of noise}
    \label{fig:noise}
\end{figure}

\begin{figure}[hb]
    \centering
    \includegraphics[width=.45\textwidth]{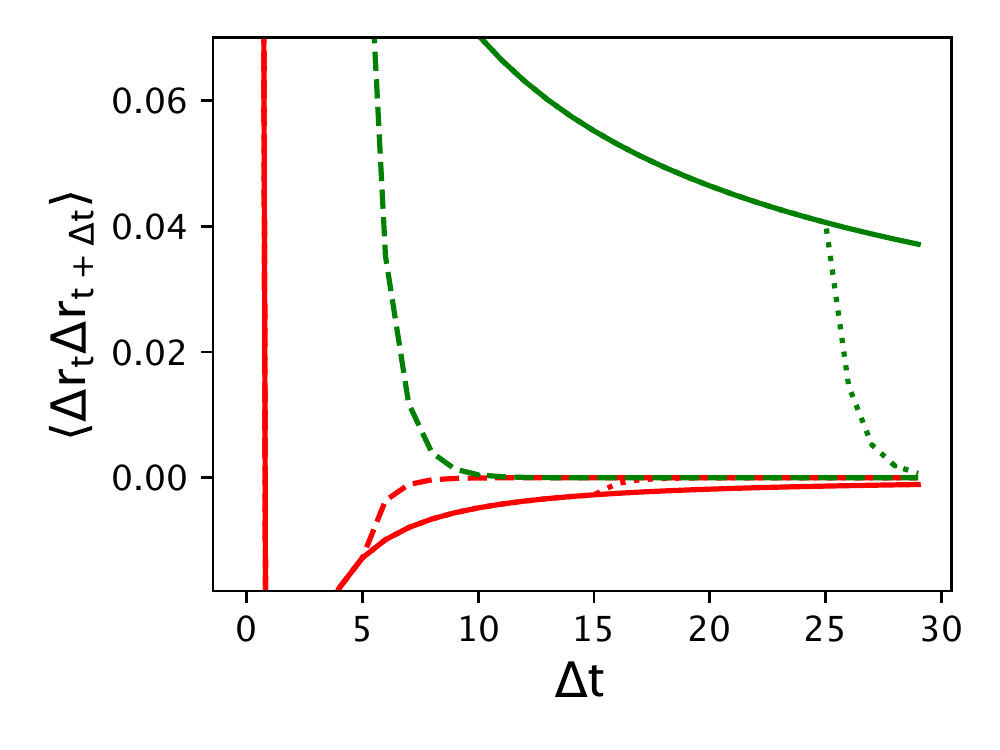}
    \caption{{\bf Temporal correlations of fBm with finite decorrelation time.}
    Autocovariance of increments of the fBm trajectory, with finite correlation time $\tau_c$, in the sub-diffusive and super-diffusive case. Red curves correspond to $\alpha = 0.6$, and $\tau_c = 5$ (dashed line), $15$ (dotted line), $\infty$ (plain line). Green curves correspond to $\alpha = 1.4$, and $\tau_c = 5$, $25$, $\infty$.}
    \label{fig:correlation}
\end{figure}

\end{document}